\pdfoutput=1

\documentclass[11pt]{article}

\usepackage[]{acl}

\usepackage{microtype}
\usepackage{graphicx}
\usepackage{subfigure}
\usepackage{booktabs} 
\usepackage{multirow}
\usepackage{stmaryrd,scalerel}
\usepackage{adjustbox}
\usepackage{alltt}

\usepackage{amsfonts}
\let\oldemptyset\emptyset

\usepackage{times}
\usepackage{latexsym}
\usepackage{xfrac}
\usepackage{xcolor}
\usepackage{tcolorbox}
\usepackage{multicol}
\usepackage{tabularx}

\tcbset{
  colframe=gray!70,         
  colback=gray!10,          
  coltitle=black,           
  fonttitle=\bfseries,      
  boxrule=0.5mm,
  top=2mm, bottom=2mm, left=2mm, right=2mm,
  sharp corners,
  width=\textwidth
}

\usepackage[T1]{fontenc}

\usepackage[utf8]{inputenc}

\usepackage{amsmath}
\usepackage{amssymb}
\usepackage{mathtools}
\usepackage{amsthm}
\usepackage{algpseudocode}
\usepackage{inconsolata}
\usepackage{bbm}
\usepackage{svg}
\usepackage[capitalize]{cleveref}
\crefname{section}{\S}{\S\S}
\crefname{table}{Tab.}{Tabs.}
\crefname{figure}{Fig.}{Figs.}

\crefname{algorithm}{Algorithm}{Algorithms}
\crefname{equation}{Eq.}{Eqs.}
\crefname{example}{Example}{Examples}
\crefname{fact}{Fact}{Facts}
\crefname{appendix}{Appendix}{Appendices}
\crefname{theorem}{Theorem}{Theorems}
\crefname{reTheorem}{Theorem}{Theorems}
\crefname{aquestion}{Question}{Questions}
\crefname{assumption}{Assumption}{Assumptions}
\crefname{lemma}{Lemma}{Lemmas}
\crefname{reLemma}{Lemma}{Lemmas}
\crefname{proposition}{Proposition}{Propositions}
\crefname{chapter}{Chapter}{Chapters}
\crefname{line}{line}{lines}
\crefname{principle}{Principle}{Principles}
\crefname{definition}{Definition}{Definitions}
\crefname{corollary}{Corollary}{Corollaries}
\crefname{Exercise}{Exercise}{Exercises}
\crefformat{section}{\S#2#1#3}

\usepackage{spverbatim}

\usepackage{macros}
\usepackage{hyphenat}
\usepackage{todonotes}
\definecolor{mintgreen}{RGB}{152, 255, 152}
\definecolor{customlightgreen}{HTML}{E0F7FA}
\makeatletter
\newcommand*\iftodonotes{\if@todonotes@disabled\expandafter\@secondoftwo\else\expandafter\@firstoftwo\fi}  %
\makeatother

\usepackage{todonotes}

\definecolor{lgreen}{HTML}{3A7D44}
\newcommand{\lgtext}[1]{\textcolor{lgreen}{#1}}
\definecolor{lblue}{HTML}{43a2ca}
\newcommand{\lbtext}[1]{\textcolor{lblue}{#1}}

\usepackage{thm-restate}

\usepackage{mathtools}

\newcommand{\defn}[1]{\textbf{#1}}
\newcommand{\word}[1]{“\textit{#1}”}

\title{A Practical Method for Generating String Counterfactuals}

 \author{Matan Avitan\textsuperscript{\normalfont1}\quad Ryan Cotterell\textsuperscript{\normalfont2} \quad  Yoav Goldberg\textsuperscript{\normalfont1,3}\quad Shauli Ravfogel\textsuperscript{\normalfont4} \\
\textsuperscript{1}Bar-Ilan University \quad \textsuperscript{2}ETH Zurich \\ \textsuperscript{3}Allen Institute for Artificial Intelligence \quad \textsuperscript{4} New York University \\ 
  { \tt\{\href{mailto:matan13av@gmail.com}{matan13av} \quad \href{mailto:shauli.ravfogel@gmail.com}{shauli.ravfogel} \quad \href{mailto:yoav.goldberg@gmail.com}{yoav.goldberg}\}@gmail.com} \quad \\\tt{\href{mailto:ryan.cotterell@inf.ethz.ch}{ryan.cotterell@inf.ethz.ch}}
   }
   
\begin{document}
\maketitle
\begin{abstract}
Interventions performed on the representation space of language models have emerged as an effective means to influence model behavior. Such methods are employed, for example, to eliminate or alter the encoding of demographic information, such as gender, within the model’s representations and, in so doing, create a counterfactual representation. However, because the intervention operates within the representation space, understanding precisely what aspects of the text it modifies poses a challenge. In this paper, we present a method to convert representation counterfactuals into string counterfactuals. We demonstrate that this approach enables us to analyze the linguistic alterations corresponding to a given representation space intervention and to interpret the features utilized to encode a specific concept. Moreover, the resulting counterfactuals can be used to mitigate bias in classification through data augmentation.

{
\vspace{0.5em}
\hspace{.5em}\includegraphics[width=1.25em,height=1.15em]{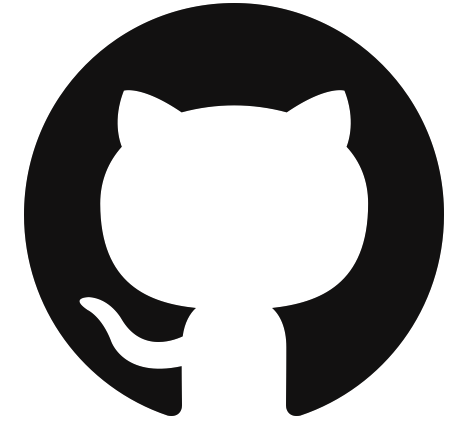}\hspace{.75em}
\parbox{\dimexpr\linewidth-7\fboxsep-7\fboxrule}{\url{https://github.com/MatanAvitan/rep-to-string-counterfactuals}}
\vspace{-.5em}%
}
\end{abstract}

\section{Introduction}\label{sec:introduction}
Interventions performed in the representation space of language models (LMs), generally $\RD$, have proven effective in understanding and exerting control over neural language models \citep{ravfogel2020null, ravfogel2021counterfactual, geva2021transformer, elazar2021amnesic, ravfogel2022adversarial, ravfogel-etal-2023-linear, belrose2023leace,li2023inference}. One popular set of techniques \defn{erases} the linear subspace associated with a human-interpretable concept $c$, e.g., \concept{gender} or \concept{sentiment}. Another widely used approach is to \defn{steer} representations from one class to another, e.g., shifting them toward a region in the representation space associated with a different class $c'$ \cite{subramani2022extracting, li2023inference, ravfogel2021counterfactual, singh2024mimic}. For instance, they could steer a representation into a region associated with negative sentiment, thereby creating \emph{counterfactual representations}. In this paper, we propose a technique to generate strings that correspond to representation-level counterfactuals, which we denote as \emph{string counterfactuals}.

\begin{figure}[t]
\centering
\begin{tcolorbox}[
    width=0.95\linewidth,
    colframe=gray!70,
    colback=gray!10,
    boxrule=0.5pt,
    titlerule=0pt,
    fonttitle=\bfseries,
    title=Original Biography
]
\footnotesize
Providing legal representation in Florida for a variety of different issues,
Barry Wax was selected to Super Lawyers for 2017--2018.
\lgtext{He} is admitted to practice before the courts in Florida.
\end{tcolorbox}


\begin{tcolorbox}[
    width=0.95\linewidth,
    colframe=gray!70,
    colback=gray!10,
    boxrule=0.5pt,
    titlerule=0pt,
    fonttitle=\bfseries,
    title=\mimic\ \mtof\
]
\footnotesize
In 2016, \lgtext{Ms.} Wax was selected by 
\lgtext{her} peers to be selected by Florida Super Lawyers 
as a Florida Super Lawyer.
\lgtext{She} represents clients in a variety of practice areas, 
including labor and employment, real estate, bankruptcy, and 
\lbtext{family law}\dots
\end{tcolorbox}


\begin{tcolorbox}[
    width=0.95\linewidth,
    colframe=gray!70,
    colback=gray!10,
    boxrule=0.5pt,
    titlerule=0pt,
    fonttitle=\bfseries,
    title={\leace\ \mtoe}
]
\footnotesize
In 2018, Barry Wax was selected as a Super Lawyer in Florida.
\lgtext{His} practice focuses on providing legal representation to clients 
in a variety of practice areas...
\end{tcolorbox}


\begin{tcolorbox}[
    width=0.95\linewidth,
    colframe=gray!70,
    colback=gray!10,
    boxrule=0.5pt,
    titlerule=0pt,
    fonttitle=\bfseries,
    title=\mimicplus\ \mtof\
]
\footnotesize
In 2016, \lgtext{Ms.} Barry was selected by 
\lgtext{her} peers to be selected by 
the Florida Super Lawyers.
\lgtext{She} represents clients in all 
\lbtext{areas of family law}, 
including but not limited to: \lbtext{legal malpractice, spousal care, child custody, and legal malpractice}\dots
\end{tcolorbox}

\caption{\footnotesize
The \emph{counterfactual lens} induces diverse string counterfactuals by leveraging different \emph{representation surgery} (i.e., representation-level interventions.)
\lgtext{Green} denotes the \emph{intended} or \emph{expected} behavior following a gender shift, while \lbtext{blue} marks \emph{stereotypical} or otherwise undesired expansions.
\vspace{-15pt}
}

\label{fig:demonstration-sample}
\end{figure}


Collectively, we refer to representation space intervention techniques as \defn{representation surgery} because they (surgically) intervene in the encoding of a concept within the representation while keeping the rest of the representation as similar as possible. In this sense, representation surgery resembles a causal intervention \cite{vig2020causal, geiger2021causal, feder2021causalm, pmlr-v162-geiger22a, guerner2023geometric, lemberger2024explaining}, and we will informally use causal language throughout the paper, referring to such modifications in the representation space as interventions. In notation, we write $\intervene_{c \rightarrow c'} \colon \RD \rightarrow \RD$ for a function that performs such an intervention.

While representation surgery techniques can create counterfactual variants of the original representations, they do not produce them at the level of natural language text. In this work, we tackle the problem of generating the counterfactual \emph{string} that corresponds to a specific representation intervention. Despite the abundance of research on representation surgery, translating such interventions into string counterfactuals remains understudied. We refer to this process as a \defn{counterfactual lens}, as it allows us to interpret representation-space counterfactuals in natural language, similar to representation-level interpretability techniques \cite{meng2022locating, nostalgebraist, belrose2023eliciting, ghandeharioun2024patchscope}. 
Constructing string counterfactuals serves various practical purposes. First, it offers a method of \defn{meta-interpretability}, aiding in the interpretation of commonly used representational intervention techniques, which themselves are often employed for interpretability. By mapping representational interventions back to the string, we can observe the lexical and higher-level semantic shifts triggered by the intervention. Second, string counterfactuals are a natural choice for data augmentation. Indeed, we demonstrate their potential to address fairness concerns in a real-world classification problem.\looseness=-1

We follow \citet{morris2023text} in developing an approach for generating string counterfactuals from representation interventions.
Let $\alphabet$ be an alphabet.
Consider a neural network that performs a mapping from a string $\str \in \kleene{\alphabet}$ to a representation $\rep = \enc(\str) \in \RD$.
\citet{morris2023text} propose an iterative algorithm to 
approximate the inverse function $\inv \colon \RD \rightarrow \kleene{\alphabet}$.
We exploit the \citeposs{morris2023text} algorithm to construct a \emph{string counterfactual} corresponding to a surgical intervention in the representation space.
Using the notation introduced so far, we are interested in computing $\str' = \inv(\intervene_{c \rightarrow c'}(\enc(\str)))$.
To the extent that $\inv$ constitutes a suitable inverse, we expect $\str'$ to be a minimally different version of $\str$ that reflects the difference between $\rep$ and $\rep'$ reflected in the representation space.

We perform experiments on a dataset of short biographies annotated with gender and profession \cite{de2019bias}. We find that swapping gender in the representation space and then generating the inverse is an effective method for producing string counterfactuals. The resulting counterfactuals exhibit some degree of gender bias, for example, a tendency to include more profession-related words in male biographies, suggesting that LMs encode subtle cues correlated with gender beyond pronouns (\cref{sec:analysis}). We further show that these counterfactuals can be used for data augmentation to improve fairness in a multiclass classification task (\cref{sec:bias}): specifically, classifiers trained on both original and counterfactual biographies (with respect to gender) exhibit reduced gender bias compared to those trained solely on the original data.

{
\begin{figure}

\centering
\usetikzlibrary{arrows.meta, calc, positioning, shapes, matrix}
\definecolor{femaleColor}{HTML}{FBB13C}   
\definecolor{maleColor}{HTML}{6CB4EE}     
\definecolor{conceptColor}{HTML}{5ab4ac}  
\definecolor{boxLine}{HTML}{333333}       
\begin{tikzpicture}[
  font=\footnotesize,
  node distance=0.5cm,
  >=Stealth,
  every node/.style={align=center}
]
\tikzset{
  boxText/.style={
    rectangle,
    rounded corners=2pt,
    draw=boxLine,
    thick,
    text width=2.5cm,
    inner sep=2pt,
    font=\footnotesize,
    align=left
  }
}
\matrix (m) [column sep=1cm, row sep=1cm] {
  \node[boxText, fill=gray!30] (origText) {%
    \textit{$\phi(\str)=\textsc{F}$}\\[2pt]
    {\scriptsize
      \lgtext{She} received \lgtext{her} undergraduate degree from Emory University\\
      and \lgtext{her} J.D.\ from the University of Houston Law Center\dots
    }
  };  
  &
  \node (hvec) {\(\mathbf{h} = \begin{pmatrix} -30 \\ 7 \\ 22 \\ 16 \end{pmatrix}\)};  
  \\
  \node (hvecprime) {\(\mathbf{h}' = \begin{pmatrix} 13 \\ -1 \\ -9 \\ 32 \end{pmatrix}\)};  
  &
  \node[boxText, fill=gray!10] (cfText) {%
    \textit{$\phi(\str')=\textsc{M}$}\\[2pt]
    {\scriptsize
      \lbtext{He} received \lbtext{his} undergraduate degree from Houston\\
      College of Law and \lbtext{his} J.D.\ degree from Emory University\dots
    }
  };  
  \\
};
\draw[->, thick] (origText.east) -- 
  node[above, font=\scriptsize]{Encoder} 
  (hvec.west);
\draw[->, thick, color=black] (hvec.south) -- 
  node[midway, sloped, above, font=\scriptsize]{Representation Surgery\\(F$\to$M)} 
  (hvecprime.north);
\draw[->, thick] (hvecprime.east) -- 
  node[above, font=\scriptsize]{Vec2Text} 
  (cfText.west);
\end{tikzpicture}

\caption{\footnotesize
An illustration of our method.
We first encode the original text to obtain a representation 
$\mathbf{h} \in \RD$.
We then apply some form of representation surgery, i.e., to steer or erase a particular concept to produce
a modified representation $\mathbf{h}'$. 
Finally, we invert the representation-level counterfactual to obtain a string-level counterfactual.\looseness=-1
}
\label{fig:intervention-lens}
\end{figure}
}

\section{Representation Surgery}
We provide a more in-depth overview of representation surgery.
Many neural networks for natural language processing construct a function $\enc \colon \kleene{\alphabet} \rightarrow \RD$ that maps a string of words over $\alphabet$, e.g., a natural language text, to a real-valued representation in $\RD$.
We call such functions \defn{language encoders} \citep{chan2024affine}.
In \Cref{sec:introduction}, we introduced a function $\intervene \colon \RD \rightarrow \RD$ that performs the intervention in the representation space. 
We consider three types of representation interventions, each discussed in a labeled paragraph below.
First, however, we will introduce some general notation.\looseness=-1

\paragraph{Notation.}
Let $p$ be a language model,\footnote{In this text, $p$ is fully decoupled from the language encoder $\enc$.
For instance, our notation allows for $p$ to some approximation to or the actual human language model (to the extent one believes in the human language model as a construct). 
However, we also allow $p$ to be deeply related to $\enc$. 
For instance, in an autoregressive language model, $\enc$ could be produced by the representation of $\eos$.} i.e., a distribution over $\kleene{\alphabet}$, let $\enc \colon \kleene{\alphabet} \rightarrow \RD$ be a language encoder. Let $\concepts=\{0,1\}$ be a binary set that stands for the different values for a concept. Binary concepts denote  whether a given property is present or not, e.g., whether or not a string $\str \in \alphabet^*$ is a biography of a man or of a woman. Furthermore, let $\phi \colon \kleene{\alphabet} \rightarrow \concepts$ be a concept encoding function.\footnote{We (simplistically) assume each string $\str$ contains exactly one concept.
Future work will relax this assumption.
}
We define the distribution\looseness=-1
\begin{equation}
  \PP(\str \mid \CC = c)  \defprop \PP(\str) \mathbbm{1} \{\phi(\str) = c\}.
\end{equation}
Then, for each $c \in \concepts$, define the following $\RD$-valued random variable
\begin{equation}
    \Xc(\str) = \enc(\str) \colon \kleene{\alphabet} \rightarrow \RD,
\end{equation}
which is distributed according to
\begin{subequations}
\begin{align}
\TrueP&(\Xc = \rep) = \mathbb{\TrueP}(\Xc^{-1}(\rep)) \\
&= \sum_{\str \in \kleene{\alphabet}}   \PP(\str \mid \CC = c) \mathbbm{1} \{ \rep = \enc(\str)\}. 
\end{align}
\end{subequations}

\paragraph{\leace{}  
 \citep{belrose2023leace}.} \leace\ is a spectral algorithm that induces log-linear guardedness \citep{ravfogel-etal-2023-linear}, i.e., it minimally (in the $L_2$ sense) modifies the $\RD$-valued random variables $\Xc$ for all $c \in \concepts$ such that there does not exist a log-linear classifier that operates at better than the accuracy of the majority class.
To achieve guardedness, \leace{} finds an oblique $D \times D$ projection matrix $\proj$ of rank $|\concepts|-1$ and a translation vector $\bias$, which are then used to define the following intervention function\looseness=-1
\begin{equation}
\leaceFunction_{\concepts \rightarrow \oldemptyset}(\Xc) \defeq \proj \Xc+ \bias.
\end{equation}

\paragraph{\mimic{} \citep{singh2024mimic}.}
\mimic, in contrast to \leace, does not merely erase the target concept from the representations, but rather takes the representations of one class (e.g., \concept{male}), and minimally modifies it such that it resembles the representations of the other class (e.g., \concept{female}). 
More precisely, it equates the first two moments of the \emph{source} class-conditional distribution to the \emph{destination} class-conditional distribution, i.e.,
\mimic{} finds a function $\mimicFunction_{c \rightarrow c'}$
such that 
\begin{subequations}
    \begin{align}
        \expectedvalue\left[ \mimicFunction_{c \rightarrow c'}(\Xc)\right] =    \expectedvalue\left[\XCTag \right] \\
        \var\left[ \mimicFunction_{c \rightarrow c'}(\Xc)\right] =   \var \left[\XCTag \right].
    \end{align}
\end{subequations}
In the case where the random variables $\Xc$ and $\XCTag$ are Gaussian distributed, MiMiC guarantees that the Wasserstein distance \cite{kantorovich1960mathematical} between $\Xc$ and $\XCTag$ is minimized.
In this case, the distance is zero.
    
\paragraph{\mimicplus.}
With \mimicplus, we further push the representations in the direction connecting the class-conditional means of the representations belonging to the two classes. 
Let $\vv \defeq \mathbb{E} \left[\mathbf{X}_{c} \right] - \mathbb{E} \left[\mathbf{X}_{c'} \right]$. 
Given a representation $\Xc(\str)$, we linearly transform 
the output of $\mimicplus$ as follows
\begin{equation}
    \mimicplusFunction_{c \rightarrow c'}(\Xc) \defeq \mimicFunction_{c \rightarrow c'}(\Xc) + \alpha \vv,
\end{equation} 
where $\alpha \geq 0$ is a scalar.
Intuitively, we move the representations towards the mean of $\Xc$.

\section{Representation Inversion}
The generative process through which natural language text is created is complex and difficult to model.
However, in some respects, it is well-approximated by modern language models. 
Concepts like gender are often conveyed subtly, and merely modifying overt indicators such as pronouns and names may not suffice  \cite{maudslay2019s}.
Instead, we leverage the fact that neural encoders capture nuanced manner in which these concepts manifest in texts.
Intervening in such representations is feasible, even \emph{without} the ability to enumerate or fully understand all linguistic features relevant to a concept.
Using representational surgery, we intervene on a concept encoded in the representation generated by an encoder
after the intervention, we apply an inverter model $\inv(\cdot)$ that maps the representation back to a string, yielding an approximate string counterfactual $\str' = \inv(\intervene(\enc(\str)))$.\looseness=-1

\paragraph{\citet{morris2023text}.}
Let $\str \in \kleene{\alphabet}$ be a sentence and let $\enc(\str)$ be its representation.
Our goal is to convert $\enc(\str)$ back into a string.
\citeposs{morris2023text} method starts by fine-tuning a language model that can be used to reconstruct an initial hypothesis $\hypothesis{0}$ of the inverse $\enc^{-1}(\str)$ given the representation $\enc(\str)$.
Then, a second language model is fine-tuned to reconstruct another hypothesis $\hypothesis{1}$ conditioned on the initial $\hypothesis{0}$, $\enc(\hypothesis{0})$, $\enc(\str)$ and the difference vector $\enc(\str) - \enc(\hypothesis{0})$.
This process is repeated $K$ times---each time $\kk \in [\KK]$, the step consists of fine-tuning the second language model conditioned on $\hypothesis{\kkSub-1}$, $\enc(\hypothesis{\kkSub-1})$, $\enc(\str)$ and the difference vector $\enc(\str) - \enc(\hypothesis{\kkSub-1})$.
The procedure ends when $\enc(\hypothesis{\kkSub})$ is sufficiently close to $\enc(\str)$ or the computational budget is exceeded. 
Then, $\hypothesis{\kkSub}$ is returned by the method as the inverse $\enc^{-1}(\str)$.
Empirically, \citet{morris2023text} find $\KK > 1$ iterations produces a more faithful inverse.

\paragraph{Putting it all together.}
Now, for a concept $c \in \concepts$ and an intervention function $\interveneCCTag$ that intervenes on that concept, we generate a counterfactual string by taking the inverse of the encoding of the string, post-intervention. Formally, the counterfactual string correspond to the following $\kleene{\alphabet}$-valued random variable:
\begin{equation}
   \XCToCTag(\str) = \enc^{-1}(\interveneCCTag (\enc(\str))),
\end{equation}
which is distributed according to
\begin{align}
&p_{c \rightarrow c'} (\str') = \mathbb{P}( \boldsymbol{R}_{c \rightarrow c'}^{-1}(\str')) \\
&\quad = \sum_{\str \in \kleene{\alphabet}} p_{c}(\str) \mathbbm{1} \{ \str' = \enc^{-1}(\interveneCCTag(\enc(\str)))\}. \nonumber
\end{align}

\section{Experimental Evaluation}\label{sec:experiments}
We now present our experimental results on gender-based interventions that modify the gender attribute in short biographical texts. We evaluate the semantic changes induced by these counterfactual interventions (\cref{sec:analysis}), assess their quality (\cref{sec:counterfactuals-quality}), and show that they help mitigate gender bias (\cref{sec:bias}).\looseness=-1

\paragraph{Inversion model.} We train a variant of the inversion model from \citet{morris2023text} on 64-token sequences and fine-tune it on the BiasBios dataset. See \cref{app:setup} for details.

\paragraph{Dataset.} 
We conduct experiments on the BiasInBios dataset \cite{de2019bias}, a large collection of short biographies sourced from the Internet. Each biography is annotated with the subject’s gender and profession.\footnote{The dataset contains 28 distinct professions.} We create natural language counterfactuals by intervening on the encoding of gender. We then use these string counterfactuals to study how gender is encoded in the LM (\cref{sec:analysis}) and to mitigate bias through data augmentation (\cref{sec:bias}).

\paragraph{Pipeline implementation.}
We trained a dedicated inversion model \citep{morris2023text} on biography representations extracted from the last layer of a GTR-base model \citep{ni2021large}, obtained by averaging word representations into a single paragraph representation. After training this inversion model, we applied one of the intervention methods to the extracted biography representations to obtain representation-level counterfactuals. For \mimic\ and \mimicplus, we set the regularization term to $10^{-5}$ and used $\alpha=2$ for \mimicplus. Finally, we applied the trained inversion model to the intervened representations to produce the desired string counterfactuals. Although the inversion model is also a GTR-base model, this is not a requirement for the method; any model could be used to create the biography representations \citep{chen-etal-2024-text}. For more details on the inversion model training setup, see \cref{app:setup}.

\begin{table}
\centering
\begin{adjustbox}{width=\columnwidth}
\begin{tabular}{lcc}
\hline
\textbf{Scenario} & \textbf{Mistral7b} & \textbf{GPT-2} \\ \hline
Original biographies & 22.62 &  104.67\\
Reconstructed biographies \\(no intervention) & 18.17 & 52.58 \\
\leace\ counterfactuals & 18.82 & 53.42 \\
\mimic\ counterfactuals & 18.29 & 51.55 \\
\mimicplus\ counterfactuals & 19.14 & 48.84 \\ \hline
\end{tabular}
\end{adjustbox}
\caption{Average perplexity for the original, reconstructed, and counterfactual biographies using the different intervention techniques generated by \texttt{Mistral7b} and \texttt{GPT-2} \citep{jiang2023mistral, radford2019language}.}
\label{tab:perplexity_results}
\end{table}

\begin{figure*}[t]
\centering
\resizebox{1.0\textwidth}{!}{ 
\begin{tcolorbox}[
    title={\textbf{Words with the Large Change in PMI}}, 
    colframe=gray!70, 
    colback=gray!10, 
    fonttitle={\large \bfseries}, 
]
    \centering
    \begin{tabularx}{\textwidth}{X X X} 
        & \textbf{Increased} & \textbf{Decreased} \\
        \hline
        \textbf{\mimic\ \mtof} & \word{ms}, \word{she's}, \word{bri}, \word{marie}, \word{mrs}, \word{girl}, \word{herself}, \word{jennifer}, \word{nicole}, \word{domestic}, \word{anne}, \word{nancy}, \word{maternal} & \word{et}, \word{himself}, \word{kau}, \word{enterprise}, \word{prof}, \word{anthony}, \word{edward}, \word{iot}, \word{acoustic}, \word{days}, \word{hardware}, \word{late} \\
        \hline
        \textbf{\mimic\ \ftom} & \word{mr}, \word{him}, \word{he's}, \word{himself}, \word{developer}, \word{chris}, \word{robert}, \word{veterinary}, \word{stephen} & \word{she's}, \word{mrs}, \word{girl}, \word{herself}, \word{nicole}, \word{female}, \word{desire}, \word{abuse}, \word{lingerie} \\
        \hline
        \textbf{\leace\ \ftoe} & \word{him}, \word{he's}, \word{mr}, \word{plays}, \word{showcase}, \word{authority}, \word{pleasure}, \word{watch}, \word{adventure} & \word{clutter}, \word{uncomfortable}, \word{classrooms}, \word{compassion}, \word{experiencing}, \word{participant}, \word{babies}, \word{engaging} \\
        \hline
        \textbf{\leace\ \mtoe} & \word{ms}, \word{colleagues}, \word{grace}, \word{leaders}, \word{happy}, \word{presenter}, \word{she's}, \word{advocates}, \word{teach} & \word{et}, \word{elite}, \word{kau}, \word{direction}, \word{theater}, \word{mentor}, \word{hollywood}, \word{photojournalism} \\
        \hline
        \textbf{\mimicplus\ \mtof} & \word{ms}, \word{women's}, \word{she's}, \word{marie}, \word{maternal}, \word{girl}, \word{female}, \word{nicole}, \word{elizabeth}, \word{maternity}, \word{joy} & \word{he}, \word{his}, \word{mr}, \word{him}, \word{michael}, \word{robert}, \word{daniel}, \word{charles}, \word{peter} \\
        \hline
        \textbf{\mimicplus\ \ftom} & \word{he's}, \word{mr}, \word{him}, \word{developer}, \word{daniel}, \word{robert}, \word{jeremy}, \word{adam}, \word{plays} & \word{she}, \word{her}, \word{ms}, \word{women}, \word{mary}, \word{jennifer}, \word{marie}, \word{herself}, \word{joy} \\
    \end{tabularx}
\end{tcolorbox}
} 
\caption{Words with the largest change in PMI.}
\label{fig:likelihood-change}
\end{figure*}

\subsection{Evaluating Counterfactuals Quality}
We now discuss our evaluation.

\label{sec:counterfactuals-quality}
\subsubsection{Automatic Evaluation}
To assess the quality of the generated counterfactuals, we computed the average perplexity of the resulting texts for each intervention technique. Perplexity is a standard measure of LM performance, with lower values indicating that the model finds the text more predictable and thus, to the extent we trust the language model, of higher fluency. 
As points of comparison, we also calculated perplexity for the original biographies and for the reconstructed biographies without any intervention (i.e., applying the inversion process of \citet{morris2023text} without modifications). The latter serves as a baseline for the degradation introduced by the inversion process itself.\looseness=-1

As shown in \cref{tab:perplexity_results}, reconstructed biographies (without intervention) consistently achieve lower perplexity than the original biographies, suggesting that the reconstruction process simplifies the text and makes it more predictable. Moreover, the counterfactuals generated by the three intervention methods (\leace, \mimic, and \mimicplus) show only a small increase in perplexity compared to the reconstructed biographies, indicating that the interventions introduce minimal degradation in fluency and largely preserve overall text quality.
While perplexity serves as a measure of fluency and predictability, it does not necessarily reflect nuanced shifts in meaning or style.
We thus evaluating using perplexity in conjunction with human evaluations.

\subsubsection{Human Evaluation}

We conducted human annotation experiments on Amazon Mechanical Turk (MTurk) to evaluate the quality of the counterfactuals and the effectiveness of our method, as detailed in \Cref{app:annotation}. Five annotators, all native English speakers from the US, UK, and Australia, were recruited and compensated for their time. They were asked to assess three aspects of the generated texts: (1) readability, (2) grammatical correctness, and (3) gender specification of the subject entity. The first two aspects measure the \emph{quality} of the counterfactual strings, while the third measures their \emph{correctness}, i.e., whether we successfully intervened in the concept of interest.\looseness=-1

For tasks (1) and (2), annotators were presented with pairs of texts (original and counterfactual) and asked to compare them in terms of readability and grammatical correctness, indicating which text was superior or whether they were comparable. For task (3), they determined the gender of the subject entity in each text (male, female, or unclear).

\paragraph{Quality.}
We performed statistical testing to evaluate whether the interventions had a significant effect on the annotators' responses regarding readability and grammatical correctness. 
The results are summarized in \cref{app:annotation} (\cref{tab:test_results}) based on \cref{tab:readability_results} and \cref{tab:grammar_results}. 
For most interventions (\leace, \mimic, and \mimicplus), the $p$-values from the one-tailed binomial tests for readability and grammar were greater than 0.05, indicating evidence for a preference for the original text over the counterfactuals. 
This suggests that our method did not degrade the quality of the text in terms of readability and grammar. 
However, for the \mimic\ \ftom\ and \mimicplus\ \ftom\ interventions, the $p$-values for readability were less than 0.05 ($p = 1.60 \times 10^{-5}$ and $p = 9.05 \times 10^{-8}$, respectively), indicating that the original text was preferred over the counterfactual in terms of readability. This suggests that the interventions did cause a degradation in readability when intervening on the perceived gender of the person described in the biography.

\paragraph{Correctness.} 
To determine whether the interventions effectively changed how annotators perceived gender, we performed chi-square tests on annotators' gender specification responses. 
Rejecting the null hypothesis under a chi-square test gives evidence that the distribution of gender identifications depends on the intervention, implying that the intervention successfully influenced the perceived gender of the text. As shown in \cref{tab:test_results}, the $p$-values for all interventions were extremely low (well below 0.05).
For instance, in the \mimic\ \ftom\ intervention, originally female biographies were annotated as male 82\% of the time after the intervention, compared to 3\% male in the original texts. This shift corresponds to a chi-square statistic of 130.56 ($p$-value $4.45 \times 10^{-29}$). Similarly, in the \mimic\ \mtof\ intervention, originally male biographies were annotated as female 86\% of the time post-intervention, compared to 10\% female in the originals, with a chi-square statistic of 131.44 ($p$-value $2.87 \times 10^{-29}$). These results show that annotators generally agreed with the intended gender changes, confirming that the interventions were effective. By contrast, \leace, an erasure method, produced more mixed outcomes. For example, when applied to originally female biographies, the proportion perceived as male rose from 11\% to 66\%, those perceived as female decreased from 85\% to 26\%, and 8\% were labeled as unclear. This pattern reflects its function as an erasure technique rather than a steering approach; see \cref{fig:demonstration-sample} and \cref{app:sample}.\looseness=-1

\subsection{Semantic Changes in the Counterfactuals}
\label{sec:analysis}
In the previous section, we validated the semantic coherence and correctness of the counterfactuals. In this section, we analyze the specific changes incurred in the inversion process. 
This analysis is performed over sentences from the dev set of BiasBios dataset whose lengths are 64 tokens or less: 7,578 biographies in the \mtofA\ direction and 6,982 biographies in the \ftomA\ direction. 

\paragraph{Pointwise mutual information.}

To quantitatively evaluate local changes induced by the counterfactual generation process, we analyze the words whose probabilities change the most between the original and counterfactual sentences.
Let $c, c' \in \concepts$ be two concepts.
In the case of concept \emph{erasure}, we may have $c' = \oldemptyset \not\in \concepts$. 
We now consider two random multisets of $M$ strings
\begin{subequations}
    \begin{align}
        S_{c \rightarrow c} &= \Lbag \str^{(m)} \mid \str \sim p_{c \rightarrow c} \Rbag _{m=1}^M \\
        S_{c \rightarrow c'} &= \Lbag \str^{(m)} \mid \str \sim p_{c \rightarrow c'} \Rbag_{m=1}^M.
    \end{align}
\end{subequations}
Then, we define a unigram distribution over $\alphabet$ induced from $S_{c \rightarrow c}$ and $S_{c \rightarrow c'}$ as follows
\begin{subequations}
\begin{align}
    p(w \mid c \rightarrow c) &\defprop \sum_{\str \in S_{c \rightarrow c}} \#(w, \str) \\
    p(w \mid c \rightarrow c') &\defprop \sum_{\str \in S_{c \rightarrow c'}} \#(w, \str),
\end{align}
\end{subequations}
where $\#(w, \str)$ returns how many times the word $w$ occurs in string $\str$.
Then, taking $p(c \rightarrow c) = p(c \rightarrow c') = \sfrac{1}{2}$, we define 
the pointwise mutual information (PMI) as follows
\begin{equation}
\mathrm{PMI}(w, c \rightarrow c') \defeq \log \frac{2 p(w, c \rightarrow c')}{p(w)}. 
\end{equation}
Manipulation then reveals that the difference of
two PMIs is the log odds ratio:
\begin{subequations}
\begin{align}
\mathrm{PMI}(w, &c \rightarrow c') - \mathrm{PMI}(w, c \rightarrow c) \\
&=  \log \frac{p(w, c \rightarrow c')}{p(w, c \rightarrow c)}.
\end{align}
\end{subequations}
Intuitively, the difference between two PMIs tells us the words whose frequency increased or decreased the most after the intervention, normalized by the amount of change incurred by the inversion process alone, i.e., inversion without an intervention. 
We additionally add a smoothing term of $10^{-6}$ when calculating the PMI. 
Finally, we sort the vocabulary according to the log-odds ratio, omitting words whose frequency is less than 5.\looseness=-1

\subsubsection{Results}
In this section, we analyze the changes in log ratios across different methods when manipulating gender concepts. See \cref{fig:demonstration-sample}  and \cref{app:sample} for a sample of the original and counterfactual sentences. See also \cref{fig:likelihood-change} for a subset of the words whose PMI changed the most, with the entire lists available in \cref{app:word_frequency}. We explore how words increase or decrease in likelihood when gendered concepts such as \concept{female} and \concept{male} are removed or altered, and we highlight the thematic shifts associated with these changes. For each method—\leace, \mimic, and \mimicplus—we find notable trends that reveal underlying gender associations in language models.\looseness=-1

\paragraph{Overall trends.} As anticipated, in the \ftomA\ direction, masculine pronouns and titles such as \word{he's}, \word{him}, \word{mr}, and \word{himself} experienced the most significant increase in likelihood. Conversely, in the \mtofA\ direction, the largest changes were observed with feminine pronouns and titles like \word{she's}, \word{ms}, \word{mrs}, and \word{herself}. Beyond pronouns, we find that some more subtle changes sometimes occur, reflecting biases in the dataset.
Furthermore, in the direction \mtofA\, the counterfactuals of the biographies of doctors often omit the \word{Dr.} prefix and replace it with \word{Ms}. Specific terms associated with professional and technical domains, such as \word{developer}, \word{managers}, \word{esl}, and \word{llp}, exhibited an increased frequency in the \ftomA\ direction, as we discuss below.
The counterfactuals generated by \mimicplus\ exhibit an overuse of stereotypical markers of the target gender, adding pronouns when they are not necessary or introducing new stereotypical information as depicted in \cref{fig:demonstration-sample}. 
This intervention tends to significantly modify the overall structure of the sentence. 
The inversion process is not perfect, and at times inflicts some changes to the original text, such as paraphrasing; see \cref{sec:counterfactuals-quality}.\looseness=-1

\begin{table*}
    \centering
    \resizebox{2.0\columnwidth}{!}{
    \begin{tabular}{@{}llll@{}}
    \toprule
    Setting & Accuracy $\uparrow$ & F1 $\uparrow$ & True Positive Rate Gender Gap $\downarrow$ \\
    \midrule
    Original biographies & 86.42$\pm{0.0}$& 79.63$\pm{0.01}$& 14.27$\pm{01}$\\
    Reconstructed biographies (no intervention) & 81.52$\pm{0.0}$& 72.08$\pm{0.0}$& 13.69$\pm{0.0}$\\
    Biographies without gender indication & 85.33$\pm{0.0}$& 77.75$\pm{0.0}$& 11.22$\pm{0.0}$\\ \midrule
    \addlinespace
    Original biographies + \leace\ counterfactuals & \textbf{86.59}$\pm{0.0}$& \textbf{81.8}$\pm{0.0}$& 12.95$\pm{0.0}$\\
    Original biographies + \mimic\ counterfactuals & 86.12$\pm{0.0}$& 80.92$\pm{0.0}$& 12.13$\pm{0.02}$\\
    Original biographies + \mimicplus\ counterfactuals & 85.76$\pm{0.0}$& 80.31$\pm{0.01}$& \textbf{10.59}$\pm{0.01}$\\
    \bottomrule
    \end{tabular}
    }
    \caption{Multi-class classification results from a log-linear model trained on top of \texttt{roberta-base} \citep{liu2019roberta}.}
    \label{tab:classification}
\end{table*}

\paragraph{\leace.} \leace\ aims to remove the ability to distinguish between stereotypically male and female representations. We find that post-intervention, texts which that originally focused on a woman now exhibit a decrease in words related to social engagement, care, and education. This reduction is evident from terms like \word{uncomfortable}, \word{strangers}, \word{volunteering}, and \word{babies}, which are often associated with stereotypically feminine social roles and nurturing activities. Educational and experiential terms like \word{seminars}, \word{classrooms}, and \word{participant} also show a decreased likelihood ratio, reflecting a diminished focus on stereotypically feminine educational themes.
Conversely, we observe an increase in the likelihood ratio of words associated with masculine pronouns and themes related to action, authority, and success. Words like \word{him}, \word{he's}, and \word{mr} show a rise in likelihood ratio, as well as action-oriented words such as \word{adventure}, \word{watch}, and \word{serve}, demonstrating a shift towards traditionally masculine concepts.
Opposite trends are shown when examining the outcomes of \leace\ \mtoe. These strings show a decrease in references to cultural, artistic, and professional domains. Words like \word{elite}, \word{theater}, and \word{mentor} diminish in likelihood ratio, suggesting a reduction in masculine-associated professional and artistic spheres.
Meanwhile, an increased likelihood ratio of words related to collaboration, leadership, and personal growth is observed. Words like \word{colleagues}, \word{leaders}, and \word{advocates} show a rise, reflecting themes of teamwork and leadership more commonly associated with femininity. Positive emotions and personal growth terms such as \word{grace} and \word{happy} also increase, signaling a shift toward nurturing and empathetic language.

\paragraph{\mimic.} 

For the \mimic\ method, performing the intervention \mtofA\ reveals a shift towards more stereotypically feminine references. Words like \word{ms}, \word{she's}, and \word{mrs} increase, as do female names like \word{marie}, \word{jennifer}, and \word{nicole}. Words relating to interpersonal relationships, emotions and caregiving, such as \word{happy} and \word{colleagues}, also rise in likelihood ratio.
When altering the female gender concept to male using \mimic, we observe an increase in male-specific references, with words like \word{mr}, \word{him}, \word{he's}, and \word{himself} rising in likelihood ratio. Male names, such as \word{dahl}, \word{chris}, and \word{stephen}, also become more prominent, along with professional and technical terms like \word{developer} and \word{managers}. Conversely, terms related such as \word{she's}, \word{mrs}, and \word{girl}, decrease in likelihood ratio, as well as names like \word{marie}, \word{nicole}, \word{anne}, \word{stephanie}, \word{susan} and terms from the social sphere such as  \word{inspire}, \word{uncomfortable}, \word{desire}, \word{strangers}, \word{classrooms}. This reflects a reduced focus on female-associated themes, particularly around care and emotional expression.

\paragraph{Summary.}
Across methods and gender concept manipulations, we observe clear patterns of thematic shifts. Removing or altering gender concepts in language models leads to changes in words associated with social roles, authority, and professional domains, reflecting underlying gender biases. These findings highlight the importance of understanding and addressing gender biases in language model development.

\subsubsection{Counterfactual Data Augmentation}
\label{sec:bias}
We have established that the proposed pipeline creates high-quality and relatively surgical counterfactuals. 
In this section, we make use of the counterfactuals to increase fairness in multiclass classification.
The BiasBios dataset exhibits an imbalance in the representation of men and women in various professions, leading to observed biases in the profession classifiers trained on this data \cite{de2019bias}. 
In our next experiment, we show how our generated string counterfactuals can be used for data augmentation. 
By adding counterfactual examples with the opposite gender label, we aim to mitigate the model's dependence on gender.

\paragraph{Setup.}
We represent each biography using the final-layer output from a GTR-base model \cite{ni2021large}. Next, we apply an intervention and decode the modified representation using a trained inversion model. For decoding, we employ beam search with a beam size of 4 and perform 20 correction steps using the pre-trained Natural Questions corrector from \citet{morris2023text}. This iterative process is repeated for each of the three intervention techniques: \leace, \mimic, and \mimicplus. The results are averaged over three models (see \cref{app:setup}). Finally, following previous work \cite{de2019bias}, we quantify bias as the RMS gap in true positive rates (TPR) between genders using a profession classifier \citep{de2019bias}.

\paragraph{Models.} 
We train a log-linear profession classifier on top of the language model \texttt{roberta-base} \citep{liu2019roberta} to predict the profession of the subject of the biography.
The classifiers are trained on the original biographies, the inverse of the biographies without intervention, the biographies without gender indications, such as pronouns (``Biographies without gender indication'' in \cref{tab:classification}), and the dataset that consists of the original biographies in addition to the corresponding counterfactuals created by \leace\, \mimic\ and \mimicplus\ ($\alpha=2$ for all experiments)

\paragraph{Results.} 
All results are presented in \cref{tab:classification}. 
Classifiers trained on the augmented dataset achieve lower TPR values (better fairness), even more so than classifiers trained on the biographies after the omission of overt gender markers. At the same time, the augmentation does not damage and even improves, main-task performance (profession classification), indicating that augmenting the dataset with intervention-induced string counterfactuals is a viable way to encourage the classifier to show invariance to the sensitive information (in our case, gender).\looseness=-1

\section{Conclusion}
We introduced a method for converting representation space interventions in language models into string-level counterfactuals. This approach bridges the gap between abstract representation manipulations and concrete textual changes, and allows us to derive the latter from the former. We demonstrated that the resulting counterfactuals are semantically coherent and that they surface some biases in the encoding of complex concepts such as gender. We additionally showed that the counterfactuals can assist in mitigating bias in classification through data augmentation.\looseness=-1

Our experiments highlight the potential of string counterfactuals for interpreting features used to encode concepts like demographic information, with important implications for fairness in NLP. However, the quality of counterfactuals depends on the inversion model, and our focus on binary attributes could be expanded in future work. In future work, we aim to refine the inversion process and extend the method to other interventions.

\section*{Limitations} \paragraph{Quality of the inversion model.} Our counterfactual generation method consists of two components: the intervention function $\intervene$ and the inversion model $\enc^{-1}$. We aimed to disentangle these two factors in our evaluation by comparing the inversions generated with interventions to those produced without intervention. 
However, complete disentanglement is challenging, and some of the observed changes may be due to imperfections in the inversion process rather than the intervention itself. We note that the inversion model is indeed imperfect and often introduces slight variations in the text (e.g., modifying numbers or geographical locations, or generating lexical paraphrases). These changes might be undesirable in certain use cases; however, improvements to the inversion model are orthogonal to our method.

\paragraph{Causal interventions.} Because the generative process of natural language texts is opaque, we inevitably rely on markers that people commonly associate with the property of interest (gender) in our evaluation. Future work should employ a controlled, synthetic setting to assess the extent to which the counterfactuals reflect the true causal factors associated with the concept of interest.

\paragraph{Representation of gender.} We rely on an existing dataset with binary gender labels. We acknowledge that this is a simplification, as gender is a complex, nonbinary construct.

\section*{Ethical Considerations} In all scenarios involving the potential application of automated methods in real-world contexts, we strongly recommend exercising caution and thoroughly evaluating the representativeness of the data, its alignment with real-world phenomena, and its potential adverse societal implications. Gender bias is a complex and multifaceted issue, and we view the experiments conducted in this paper as an initial exploration of strategies for mitigating the negative impacts of language models rather than a definitive solution to real-world bias challenges. As highlighted in the Limitations section, the use of binary gender labels arises from limitations in available data, and we anticipate that future research will enable more nuanced examinations of how gender, as a construct, manifests in text.\looseness=-1

\bibliography{anthology,custom}

\begin{thebibliography}{34}
\expandafter\ifx\csname natexlab\endcsname\relax\def\natexlab#1{#1}\fi

\bibitem[{Baboulin et~al.(2009)Baboulin, Buttari, Dongarra, Kurzak, Langou, Langou, Luszczek, and Tomov}]{BABOULIN20092526}
Marc Baboulin, Alfredo Buttari, Jack Dongarra, Jakub Kurzak, Julie Langou, Julien Langou, Piotr Luszczek, and Stanimire Tomov. 2009.
\newblock \href {https://doi.org/https://doi.org/10.1016/j.cpc.2008.11.005} {Accelerating scientific computations with mixed precision algorithms}.
\newblock \emph{Computer Physics Communications}, 180(12):2526--2533.

\bibitem[{Belrose et~al.(2023{\natexlab{a}})Belrose, Furman, Smith, Halawi, Ostrovsky, McKinney, Biderman, and Steinhardt}]{belrose2023eliciting}
Nora Belrose, Zach Furman, Logan Smith, Danny Halawi, Igor Ostrovsky, Lev McKinney, Stella Biderman, and Jacob Steinhardt. 2023{\natexlab{a}}.
\newblock \href {https://arxiv.org/abs/2303.08112} {Eliciting latent predictions from transformers with the tuned lens}.
\newblock \emph{arXiv preprint arXiv:2303.08112}.

\bibitem[{Belrose et~al.(2023{\natexlab{b}})Belrose, Schneider-Joseph, Ravfogel, Cotterell, Raff, and Biderman}]{belrose2023leace}
Nora Belrose, David Schneider-Joseph, Shauli Ravfogel, Ryan Cotterell, Edward Raff, and Stella Biderman. 2023{\natexlab{b}}.
\newblock \href {https://proceedings.neurips.cc/paper_files/paper/2023/file/d066d21c619d0a78c5b557fa3291a8f4-Paper-Conference.pdf} {Leace: Perfect linear concept erasure in closed form}.
\newblock In \emph{Advances in Neural Information Processing Systems}, volume~36, pages 66044--66063.

\bibitem[{Chan et~al.(2024)Chan, Boumasmoud, Svete, Ren, Guo, Jin, Ravfogel, Sachan, Sch{\"o}lkopf, El-Assady et~al.}]{chan2024affine}
Robin S.~M. Chan, Reda Boumasmoud, Anej Svete, Yuxin Ren, Qipeng Guo, Zhijing Jin, Shauli Ravfogel, Mrinmaya Sachan, Bernhard Sch{\"o}lkopf, Mennatallah El-Assady, et~al. 2024.
\newblock \href {https://arxiv.org/abs/2406.02329} {On affine homotopy between language encoders}.
\newblock In \emph{Proceedings of the 38th Conference on Neural Information Processing Systems}.

\bibitem[{Chen et~al.(2024)Chen, Lent, and Bjerva}]{chen-etal-2024-text}
Yiyi Chen, Heather Lent, and Johannes Bjerva. 2024.
\newblock \href {https://doi.org/10.18653/v1/2024.acl-long.422} {Text embedding inversion security for multilingual language models}.
\newblock In \emph{Proceedings of the 62nd Annual Meeting of the Association for Computational Linguistics (Volume 1: Long Papers)}, pages 7808--7827. Association for Computational Linguistics.

\bibitem[{De-Arteaga et~al.(2019)De-Arteaga, Romanov, Wallach, Chayes, Borgs, Chouldechova, Geyik, Kenthapadi, and Kalai}]{de2019bias}
Maria De-Arteaga, Alexey Romanov, Hanna Wallach, Jennifer Chayes, Christian Borgs, Alexandra Chouldechova, Sahin Geyik, Krishnaram Kenthapadi, and Adam~Tauman Kalai. 2019.
\newblock \href {https://doi.org/10.1145/3287560.3287572} {Bias in bios: {A} case study of semantic representation bias in a high-stakes setting}.
\newblock In \emph{Proceedings of the Conference on Fairness, Accountability, and Transparency}, page 120–128. Association for Computing Machinery.

\bibitem[{Elazar et~al.(2021)Elazar, Ravfogel, Jacovi, and Goldberg}]{elazar2021amnesic}
Yanai Elazar, Shauli Ravfogel, Alon Jacovi, and Yoav Goldberg. 2021.
\newblock \href {https://doi.org/10.1162/tacl_a_00359} {Amnesic probing: Behavioral explanation with amnesic counterfactuals}.
\newblock \emph{Transactions of the Association for Computational Linguistics}, 9:160--175.

\bibitem[{Feder et~al.(2021)Feder, Oved, Shalit, and Reichart}]{feder2021causalm}
Amir Feder, Nadav Oved, Uri Shalit, and Roi Reichart. 2021.
\newblock \href {https://doi.org/10.1162/coli_a_00404} {{C}ausa{LM}: Causal model explanation through counterfactual language models}.
\newblock \emph{Computational Linguistics}, 47(2):333--386.

\bibitem[{Fleiss(1971)}]{fleiss1971measuring}
Joseph~L. Fleiss. 1971.
\newblock \href {https://psycnet.apa.org/record/1972-05083-001} {Measuring nominal scale agreement among many raters}.
\newblock \emph{Psychological Bulletin}, 76(5):378--382.

\bibitem[{Geiger et~al.(2021)Geiger, Lu, Icard, and Potts}]{geiger2021causal}
Atticus Geiger, Hanson Lu, Thomas Icard, and Christopher Potts. 2021.
\newblock \href {https://proceedings.neurips.cc/paper_files/paper/2021/file/4f5c422f4d49a5a807eda27434231040-Paper.pdf} {Causal abstractions of neural networks}.
\newblock In \emph{Advances in Neural Information Processing Systems}, volume~34, pages 9574--9586.

\bibitem[{Geiger et~al.(2022)Geiger, Wu, Lu, Rozner, Kreiss, Icard, Goodman, and Potts}]{pmlr-v162-geiger22a}
Atticus Geiger, Zhengxuan Wu, Hanson Lu, Josh Rozner, Elisa Kreiss, Thomas Icard, Noah Goodman, and Christopher Potts. 2022.
\newblock \href {https://proceedings.mlr.press/v162/geiger22a.html} {Inducing causal structure for interpretable neural networks}.
\newblock In \emph{Proceedings of the 39th International Conference on Machine Learning}, volume 162, pages 7324--7338.

\bibitem[{Geva et~al.(2021)Geva, Schuster, Berant, and Levy}]{geva2021transformer}
Mor Geva, Roei Schuster, Jonathan Berant, and Omer Levy. 2021.
\newblock \href {https://doi.org/10.18653/v1/2021.emnlp-main.446} {Transformer feed-forward layers are key-value memories}.
\newblock In \emph{Proceedings of the 2021 Conference on Empirical Methods in Natural Language Processing}, pages 5484--5495. Association for Computational Linguistics.

\bibitem[{Ghandeharioun et~al.(2024)Ghandeharioun, Caciularu, Pearce, Dixon, and Geva}]{ghandeharioun2024patchscope}
Asma Ghandeharioun, Avi Caciularu, Adam Pearce, Lucas Dixon, and Mor Geva. 2024.
\newblock \href {https://arxiv.org/abs/2401.06102} {Patchscope: A unifying framework for inspecting hidden representations of language models}.
\newblock In \emph{Proceedings of the 41st International Conference on Machine Learning}.

\bibitem[{Guerner et~al.(2023)Guerner, Svete, Liu, Warstadt, and Cotterell}]{guerner2023geometric}
Cl{\'{e}}ment Guerner, Anej Svete, Tianyu Liu, Alexander Warstadt, and Ryan Cotterell. 2023.
\newblock \href {https://arxiv.org/abs/2307.15054} {A geometric notion of causal probing}.
\newblock \emph{arXiv preprint arXiv:2307.15054}.

\bibitem[{Jiang et~al.(2023)Jiang, Sablayrolles, Mensch, Bamford, Chaplot, Casas, Bressand, Lengyel, Lample, Saulnier et~al.}]{jiang2023mistral}
Albert~Q. Jiang, Alexandre Sablayrolles, Arthur Mensch, Chris Bamford, Devendra~Singh Chaplot, Diego de~las Casas, Florian Bressand, Gianna Lengyel, Guillaume Lample, Lucile Saulnier, et~al. 2023.
\newblock \href {https://arxiv.org/pdf/2310.06825} {Mistral 7b}.
\newblock \emph{arXiv preprint arXiv:2310.06825}.

\bibitem[{Kantorovich(1960)}]{kantorovich1960mathematical}
Leonid~V. Kantorovich. 1960.
\newblock \href {https://pubsonline.informs.org/doi/10.1287/mnsc.6.4.366} {Mathematical methods of organizing and planning production}.
\newblock \emph{Management Science}, 6(4):366--422.

\bibitem[{Kwiatkowski et~al.(2019)Kwiatkowski, Palomaki, Redfield, Collins, Parikh, Alberti, Epstein, Polosukhin, Devlin, Lee, Toutanova, Jones, Kelcey, Chang, Dai, Uszkoreit, Le, and Petrov}]{kwiatkowski2019natural}
Tom Kwiatkowski, Jennimaria Palomaki, Olivia Redfield, Michael Collins, Ankur Parikh, Chris Alberti, Danielle Epstein, Illia Polosukhin, Jacob Devlin, Kenton Lee, Kristina Toutanova, Llion Jones, Matthew Kelcey, Ming-Wei Chang, Andrew~M. Dai, Jakob Uszkoreit, Quoc Le, and Slav Petrov. 2019.
\newblock \href {https://doi.org/10.1162/tacl_a_00276} {Natural questions: A benchmark for question answering research}.
\newblock \emph{Transactions of the Association for Computational Linguistics}, 7:452--466.

\bibitem[{Lemberger and Saillenfest(2024)}]{lemberger2024explaining}
Pirmin Lemberger and Antoine Saillenfest. 2024.
\newblock \href {https://arxiv.org/abs/2402.00711} {Explaining text classifiers with counterfactual representations}.
\newblock \emph{arXiv preprint arXiv:2402.00711}.

\bibitem[{Li et~al.(2023)Li, Patel, Vi{\'e}gas, Pfister, and Wattenberg}]{li2023inference}
Kenneth Li, Oam Patel, Fernanda Vi{\'e}gas, Hanspeter Pfister, and Martin Wattenberg. 2023.
\newblock \href {https://arxiv.org/pdf/2306.03341.pdf} {Inference-time intervention: {E}liciting truthful answers from a language model}.
\newblock \emph{arXiv preprint arXiv:2306.03341}.

\bibitem[{Liu et~al.(2019)Liu, Ott, Goyal, Du, Joshi, Chen, Levy, Lewis, Zettlemoyer, and Stoyanov}]{liu2019roberta}
Yinhan Liu, Myle Ott, Naman Goyal, Jingfei Du, Mandar Joshi, Danqi Chen, Omer Levy, Mike Lewis, Luke Zettlemoyer, and Veselin Stoyanov. 2019.
\newblock \href {https://arxiv.org/abs/1907.11692} {{RoBERTa}: A robustly optimized {BERT} pretraining approach}.
\newblock \emph{arXiv preprint arXiv:1907.11692}.

\bibitem[{Maudslay et~al.(2019)Maudslay, Gonen, Cotterell, and Teufel}]{maudslay2019s}
Rowan~Hall Maudslay, Hila Gonen, Ryan Cotterell, and Simone Teufel. 2019.
\newblock \href {https://doi.org/10.18653/v1/D19-1530} {It{'}s all in the name: {M}itigating gender bias with name-based counterfactual data substitution}.
\newblock In \emph{Proceedings of the 2019 Conference on Empirical Methods in Natural Language Processing and the 9th International Joint Conference on Natural Language Processing}, pages 5267--5275. Association for Computational Linguistics.

\bibitem[{Meng et~al.(2022)Meng, Bau, Andonian, and Belinkov}]{meng2022locating}
Kevin Meng, David Bau, Alex Andonian, and Yonatan Belinkov. 2022.
\newblock \href {https://arxiv.org/abs/2202.05262} {Locating and editing factual associations in gpt}.
\newblock \emph{Advances in Neural Information Processing Systems}, 35:17359--17372.

\bibitem[{Micikevicius et~al.(2017)Micikevicius, Narang, Alben, Diamos, Elsen, Garcia, Ginsburg, Houston, Kuchaiev, Venkatesh et~al.}]{micikevicius2017mixed}
Paulius Micikevicius, Sharan Narang, Jonah Alben, Gregory Diamos, Erich Elsen, David Garcia, Boris Ginsburg, Michael Houston, Oleksii Kuchaiev, Ganesh Venkatesh, et~al. 2017.
\newblock \href {https://arxiv.org/abs/1710.03740} {Mixed precision training}.
\newblock \emph{arXiv preprint arXiv:1710.03740}.

\bibitem[{Morris et~al.(2023)Morris, Kuleshov, Shmatikov, and Rush}]{morris2023text}
John Morris, Volodymyr Kuleshov, Vitaly Shmatikov, and Alexander Rush. 2023.
\newblock \href {https://doi.org/10.18653/v1/2023.emnlp-main.765} {Text embeddings reveal (almost) as much as text}.
\newblock In \emph{Proceedings of the 2023 Conference on Empirical Methods in Natural Language Processing}, pages 12448--12460. Association for Computational Linguistics.

\bibitem[{Ni et~al.(2022)Ni, Qu, Lu, Dai, Hernandez~Abrego, Ma, Zhao, Luan, Hall, Chang, and Yang}]{ni2021large}
Jianmo Ni, Chen Qu, Jing Lu, Zhuyun Dai, Gustavo Hernandez~Abrego, Ji~Ma, Vincent Zhao, Yi~Luan, Keith Hall, Ming-Wei Chang, and Yinfei Yang. 2022.
\newblock \href {https://doi.org/10.18653/v1/2022.emnlp-main.669} {Large dual encoders are generalizable retrievers}.
\newblock In \emph{Proceedings of the 2022 Conference on Empirical Methods in Natural Language Processing}, pages 9844--9855. Association for Computational Linguistics.

\bibitem[{nostalgebraist(2020)}]{nostalgebraist}
nostalgebraist. 2020.
\newblock \href {https://www.lesswrong.com/posts/AcKRB8wDpdaN6v6ru/interpreting-gpt-the-logit-lens} {Interpreting {GPT}: The logit lens}.

\bibitem[{Radford et~al.(2019)Radford, Wu, Child, Luan, Amodei, and Sutskever}]{radford2019language}
Alec Radford, Jeff Wu, Rewon Child, David Luan, Dario Amodei, and Ilya Sutskever. 2019.
\newblock \href {https://api.semanticscholar.org/CorpusID:160025533} {Language models are unsupervised multitask learners}.

\bibitem[{Ravfogel et~al.(2020)Ravfogel, Elazar, Gonen, Twiton, and Goldberg}]{ravfogel2020null}
Shauli Ravfogel, Yanai Elazar, Hila Gonen, Michael Twiton, and Yoav Goldberg. 2020.
\newblock \href {https://doi.org/10.18653/v1/2020.acl-main.647} {Null it out: Guarding protected attributes by iterative nullspace projection}.
\newblock In \emph{Proceedings of the 58th Annual Meeting of the Association for Computational Linguistics}, pages 7237--7256. Association for Computational Linguistics.

\bibitem[{Ravfogel et~al.(2023)Ravfogel, Goldberg, and Cotterell}]{ravfogel-etal-2023-linear}
Shauli Ravfogel, Yoav Goldberg, and Ryan Cotterell. 2023.
\newblock \href {https://doi.org/10.18653/v1/2023.acl-long.523} {Log-linear guardedness and its implications}.
\newblock In \emph{Proceedings of the 61st Annual Meeting of the Association for Computational Linguistics (Volume 1: Long Papers)}, pages 9413--9431. Association for Computational Linguistics.

\bibitem[{Ravfogel et~al.(2021)Ravfogel, Prasad, Linzen, and Goldberg}]{ravfogel2021counterfactual}
Shauli Ravfogel, Grusha Prasad, Tal Linzen, and Yoav Goldberg. 2021.
\newblock \href {https://doi.org/10.18653/v1/2021.conll-1.15} {Counterfactual interventions reveal the causal effect of relative clause representations on agreement prediction}.
\newblock In \emph{Proceedings of the 25th Conference on Computational Natural Language Learning}, pages 194--209. Association for Computational Linguistics.

\bibitem[{Ravfogel et~al.(2022)Ravfogel, Vargas, Goldberg, and Cotterell}]{ravfogel2022adversarial}
Shauli Ravfogel, Francisco Vargas, Yoav Goldberg, and Ryan Cotterell. 2022.
\newblock \href {https://doi.org/10.18653/v1/2022.emnlp-main.405} {Kernelized concept erasure}.
\newblock In \emph{Proceedings of the 2022 Conference on Empirical Methods in Natural Language Processing}, pages 6034--6055. Association for Computational Linguistics.

\bibitem[{Singh et~al.(2024)Singh, Ravfogel, Herzig, Aharoni, Cotterell, and Kumaraguru}]{singh2024mimic}
Shashwat Singh, Shauli Ravfogel, Jonathan Herzig, Roee Aharoni, Ryan Cotterell, and Ponnurangam Kumaraguru. 2024.
\newblock \href {https://arxiv.org/abs/2402.09631} {{MiMiC}: {M}inimally modified counterfactuals in the representation space}.
\newblock \emph{arXiv preprint arXiv:2402.09631}.

\bibitem[{Subramani et~al.(2022)Subramani, Suresh, and Peters}]{subramani2022extracting}
Nishant Subramani, Nivedita Suresh, and Matthew Peters. 2022.
\newblock \href {https://doi.org/10.18653/v1/2022.findings-acl.48} {Extracting latent steering vectors from pretrained language models}.
\newblock In \emph{Findings of the Association for Computational Linguistics: ACL 2022}, pages 566--581. Association for Computational Linguistics.

\bibitem[{Vig et~al.(2020)Vig, Gehrmann, Belinkov, Qian, Nevo, Singer, and Shieber}]{vig2020causal}
Jesse Vig, Sebastian Gehrmann, Yonatan Belinkov, Sharon Qian, Daniel Nevo, Yaron Singer, and Stuart~M. Shieber. 2020.
\newblock \href {https://arxiv.org/abs/2004.12265} {Causal mediation analysis for interpreting neural {NLP}: {T}he case of gender bias}.
\newblock \emph{arXiv preprint arXiv:2004.12265}.

\end{thebibliography}
\bibliographystyle{acl_natbib}

\newpage
\appendix
\onecolumn
\section*{Appendix}

\section{Experimental setup}
\label{app:setup}

\paragraph{Training an inversion model.} \citet{morris2023text} introduced an approach for converting representations into strings. To effectively invert the representations derived from the BiasBios dataset, we trained a dedicated inversion model on 64-token sequences from the Natural Questions dataset \cite{kwiatkowski2019natural}. This decision was informed by the observation that the median biography length in the BiasBios dataset is 72 tokens. The model architecture is GTR-base \citep{ni2021large}, as originally used in vec2text \citep{morris2023text}. The inversion process consists of two components: the inversion model and a corrector model (both are GTR-base LMs). Empirical results demonstrate that training both components improves the quality of the reconstructed text. The training procedure involved training the inversion model for 30 epochs on the Natural Questions dataset \citep{kwiatkowski2019natural} with a batch size of 4096, followed by fine-tuning for an additional 20 epochs on the BiasBios dataset \citep{de2019bias} with a batch size of 512. Subsequently, the corrector model was trained on the BiasBios dataset for 10 epochs using a batch size of 128 samples.

\paragraph{Training profession classifiers.} To quantify the causal effect of counterfactuals on predicting an individual's profession, we utilized \texttt{roberta-base} \citep{liu2019roberta} classifiers trained on both the counterfactuals and the corresponding original biographies, as outlined in \cref{tab:classification}. Each classifier was trained with three different seeds, and we report the mean and standard deviation of the metrics obtained from the checkpoint with the lowest validation loss for each seed. The classifiers were trained for 10 epochs on the entire BiasBios biography dataset, with sequences truncated to 64 tokens. This dataset comprises 7,578 male biographies and 6,982 female biographies. For each original sample, its corresponding counterfactual was included in the training set. We used a batch size of 1024 samples for training and 4096 for evaluation. Furthermore, 6\% of the samples were used for learning rate warm-up, with an initial learning rate set to $2\times10^{-5}$. We also employed half-precision quantization (fp16) for the network's weights \citep{BABOULIN20092526,micikevicius2017mixed}. The results reported in \cref{tab:classification} were calculated on the entire BiasBios development set (39,369 samples), with sequences truncated to 64 tokens.
\section{Word PMI Analysis}
\label{app:word_frequency}

We provide the words most changed due to \mimic\ , \leace\ and \mimicplus\ interventions below. \\
\subsection{\mimic}
\begin{itemize}
    \item words whose likelihood most decreased in direction \mtofA\ :
    
[\word{et}, \word{himself}, \word{kau}, \word{enterprise}, \word{really}",\word{prof}, \word{anthony}, \word{ch}, \word{edward}, \word{iot}, \word{0560}, \word{1978}, \word{acoustic}, \word{biggest}, \word{steven}, \word{founding}, \word{days}, \word{hardware}, \word{patience}, \word{late}, \word{reputed}, \word{3d}, \word{run}, \word{stephen}, \word{trustee}, \word{boy}, \word{theater}, \word{join}, \word{detection}, \word{rather}]

    \item words whose likelihood most increased in direction \mtofA\ :

[\word{ms}, \word{she's}, \word{*}, \word{bri}, \word{marie}, \word{mrs}, \word{girl}, \word{herself}", \word{jennifer}", \word{002412}, \word{nicole}, \word{women's}, \word{happy}, \word{newborn}, \word{andrea}, \word{domestic}, \word{exploring}, \word{mn}, \word{colleagues}, \word{setting}, \word{anne}, \word{elizabeth}, \word{1215242727}, \word{donna}, \word{geriatric}, \word{nancy}, \word{upon}, \word{maternal}, \word{picture}, \word{1215191916}]

    \item words whose likelihood most decreased in direction \ftomA\ :

[\word{she's}, \word{mrs}, \word{girl}, \word{marie}, \word{|}, \word{herself}, \word{clutter}", \word{inspire}, \word{uncomfortable}, \word{nicole}, \word{female}, \word{promotes}, \word{anne}", \word{desire}, \word{13}, \word{abuse}, \word{lingerie}, \word{caring}, \word{elder}, \word{strangers}, \word{classrooms}, \word{stephanie}, \word{mn}, \word{susan}, \word{refugee}, \word{runway}, \word{21}, \word{within}, \word{59}, \word{plants}]

    \item words whose likelihood most increased in direction \ftomA:

[\word{mr}, \word{him}, \word{he's}, \word{dahl}", \word{himself}, \word{1st}, \word{peers}, \word{plays}, \word{.0}, \word{2019}, \word{developer}, \word{chris}, \word{x}, \word{robert}, \word{veterinary}, \word{esl}, \word{lifetime}, \word{llp}, \word{wallpapers}, \word{adventure}, \word{chance}, \word{managers}, \word{watch}, \word{humour}, \word{murya}, \word{1003021313}, \word{stephen}, \word{list}, \word{say}, \word{concerned}]

\end{itemize}

\subsection{\leace}
\begin{itemize}
    \item words whose likelihood most decreased in direction \ftoe:
    
[\word{clutter}, \word{uncomfortable}, \word{strangers}, \word{front}, \word{classrooms},\word{volunteering}, \word{0000}, \word{never}, \word{travelling}, \word{seminars}, \word{compassion}, \word{cute}, \word{humanitarian}, \word{pre-}, \word{experimental}, \word{accredited}, \word{experiencing}, \word{partnerships}, \word{distribution}, \word{off}, \word{participant}, \word{implementing}, \word{babies}, \word{funny}, \word{die}, \word{photographing}, \word{1903021717}, \word{words}, \word{engaging}, \word{engages}]

    \item words whose likelihood most increased in direction \ftoe:

[\word{him}, \word{he's}, \word{mr}, \word{hunger}, \word{eat}, \word{himself}, \word{plays}, \word{hot}, \word{showcase}, \word{inspiring}, \word{fair}, \word{authority}, \word{1979}, \word{llp}, \word{watch}, \word{pleasure}, \word{cns}, \word{beyond}, \word{failure}, \word{per}, \word{meets}, \word{suny}, \word{adventure}, \word{agricultural}, \word{serve}, \word{greater}, \word{luxury}, \word{idea}, \word{night}, \word{reuters}]

    \item words whose likelihood most decreased in direction \mtoe:

[\word{et}, \word{elite}, \word{kau}, \word{ch}, \word{pastoral}, \word{direction}, \word{0560}, \word{choice}, \word{august}, \word{patience}, \word{cinema}, \word{restaurant}, \word{58}, \word{theater}, \word{join}, \word{rather}, \word{composing}, \word{tn}, \word{reviewer}, \word{kent}, \word{core}, \word{effect}, \word{mentor}, \word{significant}, \word{entertainment}, \word{hollywood}, \word{something}, \word{photojournalism}, \word{friend}, \word{demand}]

    \item words whose likelihood most increased in direction \mtoe:

[\word{ms}, \word{colleagues}, \word{grace}, \word{prepare}, \word{leaders}, \word{mediations}, \word{greater}, \word{setting}, \word{grown}, \word{happy}, \word{publication}, \word{writers}, \word{similar}, \word{presenter}, \word{counsels}, \word{1903021515}, \word{employee}, \word{19th}, \word{bi}, \word{she's}, \word{wilderness}, \word{bad}, \word{embedded}, \word{believer}, \word{detail}, \word{promotion}, \word{advocates}, \word{teach}, \word{mri}, \word{dedication}]
\end{itemize}

\subsection{\mimicplus}
\begin{itemize}
    \item words whose likelihood most decreased in direction \mtofA\ :
    
[\word{he}, \word{his}, \word{mr}, \word{him}, \word{he's}, \word{michael}, \word{william}, \word{et}, \word{elite}, \word{mark}, \word{andrew}, \word{robert}, \word{man}, \word{paul}, \word{brian}, \word{richard}, \word{himself}, \word{daniel}, \word{engineer}, \word{funded}, \word{alan}, \word{joseph}, \word{charles}, \word{distributed}, \word{–}, \word{peter}, \word{developer}, \word{kau}, \word{subject}, \word{adam}]

    \item words whose likelihood most increased in direction \mtofA\ :

[\word{ms}, \word{women's}, \word{she's}, \word{marie}, \word{maternal}, \word{girls}, \word{girl}, \word{1417191916}, \word{1417191997}, \word{empowerment}, \word{michelle}, \word{nicole}, \word{female}, \word{jennifer}, \word{elizabeth}, \word{mrs}, \word{nurses}, \word{parenting}, \word{mary}, \word{promotion}, \word{practitioners}, \word{birth}, \word{empowering}, \word{holistic}, \word{mom}, \word{mothers}, \word{maternity}, \word{woman's}, \word{crisis}, \word{joy}]

    \item words whose likelihood most decreased in direction \ftomA\:

[\word{she}, \word{her}, \word{ms}, \word{women}, \word{she's}, \word{mother}, \word{ki}, \word{women's}, \word{mrs}, \word{woman}, \word{elementary}, \word{january}, \word{mary}, \word{daughter}, \word{girl}, \word{jennifer}, \word{marie}, \word{|}, \word{assisting}, \word{lisa}, \word{jessica}, \word{herself}, \word{elizabeth}, \word{joy}, \word{sexual}, \word{pregnancy}, \word{amy}, \word{sexuality}, \word{opportunities}, \word{rachel}]

    \item words whose likelihood most increased in direction \ftomA:

[\word{he's}, \word{mr}, \word{him}, \word{guy}, \word{x}, \word{*}, \word{ka}, \word{himself}, \word{developer}, \word{daniel}, \word{1st}, \word{robert}, \word{1003021313}, \word{juicy}, \word{jeremy}, \word{nephrology}, \word{peers}, \word{chairman}, \word{adam}, \word{hardware}, \word{bi}, \word{matthew}, \word{mark}, \word{acoustic}, \word{//}, \word{christopher}, \word{plays}, \word{.0}, \word{player}, \word{forum}]
\end{itemize}

\section{Human Annotation}
\label{app:annotation}
We conducted human annotation experiments to evaluate the quality of the interventions using Amazon Mechanical Turk (MTurk). Five annotators, all native English speakers from the US, UK, and Australia, were recruited for this task. The annotators were compensated for their work in line with standard MTurk rates. This selection process ensured that the annotators had a high degree of fluency in English.
Annotators were required to complete three tasks: (1) assess the readability of pairs of sentences (2) assess their grammatical correctness, and (3) determine the subject entity gender for each sentence. These tasks were designed to evaluate the quality and correctness of the generated counterfactuals compared to the original biographies, following the annotation guidelines; see \cref{app:annotation_form}.
In tasks (1) and (2), annotators were presented with two texts, labeled Text A and Text B. They were asked to compare the readability and grammatical correctness of the texts, selecting which was more readable and grammatically correct, or indicating that both were comparable. In task (3), annotators were asked to identify the gender of the subject entity in the sentence: male, female, or unclear.
To analyze the results, we performed a chi-Square Test of Independence to statistically evaluate whether there was a significant difference in the annotation responses before and after applying the interventions.

\begin{table}
\centering

\begin{tabular}{p{3.5cm}|rrr}
\toprule
 Intervention & Same &  Original &  Counterfactual \\
\midrule
   {\leace}  {\ftoe} & 29 &      55 &      16 \\
   {\leace} {\mtoe} & 20 &      52 &      28 \\
   {\mimic} {\ftom} & 8 &      71 &      21 \\
   {\mimic} {\mtof} & 15 &      41 &      44 \\
   {\mimicplus} {\ftom} & 10 &      76 &      14 \\
   {\mimicplus} {\mtof} & 13 &      41 &      46 \\
\bottomrule
\end{tabular}
\caption{Readability annotation results}
\label{tab:readability_results}
\end{table}

\begin{table}[h]
\centering
\begin{tabular}{p{3.5cm}|rrr}
\toprule
 Intervention & Same &  Original &  Counterfactual \\
\midrule
  {\leace} {\ftoe} & 48 &      36 &      16 \\
  {\leace} {\mtoe} & 58 &      29 &      13 \\
   {\mimic} {\ftom} & 57 &      37 &      6 \\
   {\mimic} {\mtof} & 52 &      30 &      18 \\
   {\mimicplus} {\ftom} & 49 &      38 &      13 \\
   {\mimicplus} {\mtof} & 45 &      34 &      21 \\
\bottomrule
\end{tabular}
\caption{Grammar annotation results}
\label{tab:grammar_results}
\end{table}

\paragraph{Hypotheses.}
We formulated our hypotheses separately for each task and applied the appropriate statistical tests:

\begin{itemize} \item \textbf{Readability and Grammar (One-Tailed Binomial Test)} \begin{itemize} \item \textbf{Null Hypothesis} ($H_0$): The original text is \emph{not} preferred over the counterfactual text in terms of readability and grammar, i.e., the probability of preferring the original biography is less than or equal to 0.5. \item \textbf{Alternative Hypothesis} ($H_1$): The original text is preferred over the counterfactual text in terms of readability/grammar, i.e., the probability of preferring Text A is greater than 0.5. \end{itemize} \item \textbf{Gender Specification (Chi-Square Test of Independence)} \begin{itemize} \item \textbf{Null Hypothesis} ($H_0$): The distribution of gender identification is independent of the intervention, i.e., the intervention does not affect how annotators perceive the gender of the subject entity. \item \textbf{Alternative Hypothesis} ($H_1$): The distribution of gender identification is dependent on the intervention, i.e., the intervention affects how annotators perceive the gender of the subject entity. \end{itemize} \end{itemize}

\paragraph{Results.}

\begin{table*}[h]
    \centering
    \resizebox{0.9\textwidth}{!}{
    \begin{tabular}{@{}llll@{}}
    \toprule
    \textbf{Intervention} & \textbf{Test} & \textbf{Test Statistic} & \textbf{$p$-value} \\
    \midrule
    \multirow{3}{*}{{\leace} {\ftoe}} & Readability (Binomial Test) & $k = 55$, $n = 100$ & $p = 0.18$ \\
    & Grammar (Binomial Test) & $k = 36$, $n = 100$ & $p = 1.00$ \\
    & Gender Specification (Chi-Square Test) & $\chi^2 = 71.98$ & $p = 2.34 \times 10^{-16}$ \\
    \addlinespace
    \multirow{3}{*}{{\leace} {\mtoe}} & Readability (Binomial Test) & $k = 52$, $n = 100$ & $p = 0.38$ \\
    & Grammar (Binomial Test) & $k = 29$, $n = 100$ & $p = 1$ \\
    & Gender Specification (Chi-Square Test) & $\chi^2 = 24.87$ & $p = 3.97 \times 10^{-6}$ \\
    \addlinespace
    \multirow{3}{*}{{\mimic} {\ftom}} & Readability (Binomial Test) & $k = 71$, $n = 100$ & $p = 1.60 \times 10^{-5}$ \\
    & Grammar (Binomial Test) & $k = 37$, $n = 100$ & $p = 1.00$ \\
    & Gender Specification (Chi-Square Test) & $\chi^2 = 130.56$ & $p = 4.45 \times 10^{-29}$ \\
    \addlinespace
    \multirow{3}{*}{{\mimic} {\mtof}} & Readability (Binomial Test) & $k = 41$, $n = 100$ & $p = 0.97$ \\
    & Grammar (Binomial Test) & $k = 30$, $n = 100$ & $p = 1.00$ \\
    & Gender Specification (Chi-Square Test) & $\chi^2 = 131.44$ & $p = 2.87 \times 10^{-29}$ \\
    \addlinespace
    \multirow{3}{*}{{\mimicplus} {\ftom}} & Readability (Binomial Test) & $k = 76$, $n = 100$ & $p = 9.05 \times 10^{-8}$ \\
    & Grammar (Binomial Test) & $k = 38$, $n = 100$ & $p = 0.99$ \\
    & Gender Specification (Chi-Square Test) & $\chi^2 = 103.94$ & $p = 2.69 \times 10^{-23}$ \\
    \addlinespace
    \multirow{3}{*}{{\mimicplus} {\mtof}} & Readability (Binomial Test) & $k = 41$, $n = 100$ & $p = 0.97$ \\
    & Grammar (Binomial Test) & $k = 34$, $n = 100$ & $p = 1.00$ \\
    & Gender Specification (Chi-Square Test) & $\chi^2 = 134.58$ & $p = 5.97 \times 10^{-30}$ \\
    \bottomrule
    \end{tabular}
    }
    \caption{Test results for readability, grammar, and gender specification tasks across \leace\, \mimic\, and \mimicplus\ interventions. $p$-values above 0.05 in the binomial tests indicate no significant preference for the original text over the counterfactual, suggesting that the interventions did not degrade text quality. $p$-values below 0.05 in the chi-square tests indicate statistically significant differences in gender specification after the interventions.}
    \label{tab:test_results}
    
\end{table*}

\begin{table}[h]
\centering

\begin{tabular}{l l lll}
\toprule
\textbf{Technique} & \textbf{Data Type} & \textbf{Male} & \textbf{Female} & \textbf{Unclear} \\
\midrule
{\leace} {\ftoe}           & Original     & 11 & 85 & 4 \\
                      & Intervention & 66 & 26 & 8 \\
\hline
{\leace} {\mtoe}           & Original     & 92 & 5 & 3 \\
                      & Intervention & 64 & 32 & 4 \\
\hline
{\mimic} {\ftom}  & Original     & 3 & 97 & 0 \\
                      & Intervention & 82 & 17 & 1 \\
\hline
{\mimic} {\mtof}  & Original     & 90 & 10 & 0 \\
                      & Intervention & 9 & 86 & 5 \\
\hline
{\mimicplus} {\ftom} & Original     & 12 & 87 & 1 \\
                      & Intervention & 84 & 16 & 0 \\
\hline
{\mimicplus} {\mtof} & Original     & 91 & 9 & 0 \\
                      & Intervention & 9 & 88 & 3 \\
\bottomrule
\end{tabular}
\caption{Gender annotation results for different intervention techniques}
\label{tab:gender_results_extended}
\end{table}

The results of the statistical tests are summarized in \cref{tab:test_results}, \cref{tab:gender_results_extended} and \cref{tab:grammar_results}.
For the readability and grammar tasks, we performed one-tailed binomial tests. The number of times the original text was preferred ($k$) and the total number of observations ($n$) are reported, along with the $p$-values.
For the gender specification task, we performed chi-square tests of independence, reporting the chi-square statistic and the $p$-value.

\paragraph{Conclusions.}
\begin{itemize}
\item \textbf{Readability and Grammar:} For most interventions, the $p$-values from the one-tailed binomial tests are greater than 0.05, indicating that we fail to reject the null hypothesis. This suggests that the original text was not significantly preferred over the counterfactual in terms of readability and grammatical correctness, implying that the interventions did not degrade text quality.

However, for the \mimic\ \ftom\ and \mimicplus\ \ftom\ interventions, the $p$-values for readability are less than 0.05 ($p = 1.60 \times 10^{-5}$ and $p = 9.05 \times 10^{-8}$, respectively). This means we reject the null hypothesis in these cases, indicating that the original text was significantly preferred over the counterfactual in terms of readability. This suggests that these interventions may have led to a degradation in readability when altering gender from female to male.

\item \textbf{Gender Specification:} For all interventions, the $p$-values from the chi-square tests are significantly less than 0.05, leading us to reject the null hypothesis. This indicates that the interventions had a statistically significant effect on how annotators perceived the gender of the subject entity. Therefore, the interventions were effective in altering the perceived gender in the texts.

\end{itemize}
An agreement between the annotators was measured by Fleiss' $\kappa$ score \cite{fleiss1971measuring}. 
For task (1), comparing the readability of the sentence pairs, Fleiss' $\kappa$ was 0.23, indicating fair agreement among the annotators.
For task (2), comparing the grammar level of the sentence pairs, Fleiss' $\kappa$ was 0.21, indicating fair agreement among the annotators.
For task (3), determining the subject entity's gender, Fleiss' $\kappa$ was 0.6, indicating moderate to substantial agreement.

The counterfactual was randomly presented as Text A or Text B with a uniform distribution. Moreover the counterfactual sentence was generated by applying one of the three intervention techniques followed by the inversion model. The samples were drawn uniformly with replacement using a random sampling generator.

The exact annotation guidelines provided to the annotators are given in \Cref{app:annotation_form}.


\section{Annotation Guidelines}
\label{app:annotation_form}
\begin{verbatim}
Overview
You will be provided with two texts, labeled Text A and Text B. 
Your task is to evaluate these texts based on their:
    * Readability
    * Grammatical correctness
    * Entity gender specification

Examples
Read the following two texts (Text A and B) and answer the following questions:

Text A:
In this capacity he will assist clients in matters involving estates, trusts, wills,
guardianships, asset disputes, powers of attorney, and advanced medical directives.
Text B:
In this capacity, she will assist clients in a variety of medical matters, 
including elder care, 
medical malpractice, wills, trusts, powers of attorney, guardianships, 
and advanced medical directives.

Question 1: Which of the texts A or B is more readable and understandable? If both 
texts are comparable in terms of readability, select Same. 
Answer: Same

Question 2: Which of the texts A or B is more grammatically correct? If both 
texts are comparable in terms of grammar, select Same.
Answer: Same

Question 3: Is the subject entity male, female, or unclear? 
Answer:
    * Text A: Male
    * Text B: Female
    
Text A:
She studied at the Wimbledon School of Art 1980-84 and later on with Cecil Collins 
and Sybil Andrews. She has traveled extensively, setting up homes and painting in Kenya, 
Dubai, Canada, and Jerusalem.

Text B:
He studied at the London College of Art with Andrew Davies and Sybil Kennedy. Since 1987, 
he has traveled to New Zealand, Canada, Israel, Kenya, Australia, and New Zealand, 
where he studied a range of painting-in-residences including...

Question 1: Which of the texts A or B is more readable and understandable? If both texts 
are comparable in terms of readability, select Same.
Answer: Text A

Question 2: Which of the texts A or B is more grammatically correct? If both texts are 
comparable in terms of grammar, select Same.
Answer: Same

Readability and Grammatical Correctness
Read the following two texts (Text A and B) and answer the following questions:
Text A:
$text_a
Text B:
$text_b
Question 1: Which of the texts A or B is more readable and understandable? If both texts 
are comparable in terms of readability, select Same. 

Possible Answers:
    * Same
    * Text A
    * Text B
    
Question 2: Which of the texts A or B is more grammatically correct? If both texts are 
comparable in terms of grammar, select Same. 

Possible Answers:
    * Same
    * Text A
    * Text B
    
Gender Annotation
For each text, determine the gender of the subject entity.
Text A:
$text_a
Is the subject entity male, female, or unclear?

Possible Answers:
    * Male
    * Female
    * Unclear
Text B:
$text_b
Is the subject entity male, female, or unclear?

Possible Answers:
    * Male
    * Female
    * Unclear
\end{verbatim}

\clearpage
\newpage

\section{Intervention inversion sample}
\label{app:sample}
In \cref{tab:random-sample} we provide a random sample of the counterfactuals generated by the different methods.

\begin{table*}[h]
    \centering
    \resizebox{0.95\columnwidth}{!}{
    \begin{tabular}{p{2.5cm}p{8cm}p{8cm}}
    \toprule
         \textbf{Method} & \textbf{Inversion without intervention}  & \textbf{Intervention + Inversion} 
\\
         \hline
         \mimic\ \ftom\ & A 2017 Sports Illustrated Swimsuit Trend Hero, she talks about how she learned to embrace her body's curves, embrace their natural curves, and never let it get in the way of her career.
&A 2017 Sports Illustrated Swimsuit Hero, he embraces his flaws to become a hero, but never lets risk get in the way of his self - confidence.
\\ \hline
         \mimic\ \ftom\ & Prior to moving to New York, Naomi worked as a registered nurse for six years, specializing in cardiology. During this time, Naomi also worked in the Emergency Department at Duke University Medical Center. Naomi has a Master's Degree in Nursing. She is&Prior to joining Manhattan, she worked as an RN in the Emergency Department, specializing in cardiology. During her time at UMDNJ - NewYork, Naomi completed a Master's Degree in Cardiology.\\ \hline
         \mimic\ \mtof\ & Dr. Ahn attended medical school at Northwestern University, Feinberg School of Medicine and completed a residency at the Children's Memorial Hospital. He is board certified by the American Board of Pediatrics.&Dr. Ahn attended medical school at Northwestern University Feinberg School of Medicine and completed her residency in Pediatrics at the Children's Hospital. She is board - certified by the American Academy of Nurse Practitioners in Pediatrics.\\ \hline
         \mimic\ \mtof\ & He earned his Ph.D. in business administration from Penn State University's Smeath School of Business. His research interests include business process innovation, data mining, information security, and healthcare operations research.&She earned her doctorate in business administration from Penn State University's Smeal School of Business and her master's and bachelor's degrees from Penn State University. Her research interests include healthcare operations research, e - health, innovation, patient satisfaction, process\\ \hline
         \leace\ \ftoe & She earned Bachelor of Science degrees in Biology and Chemistry from Metropolitan State College in DeKalb, MN. Dean obtained her Master of Science degree in physician assistant studies from Denver Medical College.
&He earned a Bachelor of Science degree in Biology and Chemistry from Metropolitan State College in Denver, Colorado. Dean obtained his Master of Medical Science degree in physician assistant studies from Western Michigan University.
\\ \hline
         \leace\ \ftoe & She has devoted her life to helping people to achieve their health goals through diet, nutrition and lifestyle counseling. Diana can be contacted at : danielle.danielle.co.uk
&He has helped hundreds of people to achieve their health and lifestyle through nutrition. Diana helps people to navigate their way of life through nutrition and counseling. For more information, please visit : http://www.daniellebakhron.com
\\ \hline
         \leace\ \mtoe & He focuses his practice on understanding how to protect data and privacy across the entire enterprise. Mr. Robinson also works with federal and state agencies to develop data security standards, best practices, cybersecurity best practices, and best practices to protect data and privacy across the entire enterprise.
&She focuses her practice on partnering with businesses, government agencies, and private companies to develop and implement best practices in data security, data privacy, and cybersecurity. Mr. Robinson also counsels businesses on how to protect their data assets and comply with federal data security standards.
\\ \hline
         \leace\ \mtoe & He graduated with a degree in Architecture and Design from the Architectural Association School of Design and Design. His architectural practice is based in Notting Hill London.
& He graduated from the Architectural Association School of Architecture and Design based in Notting Hill, London. He also has a degree in architecture and design. His practice is based in notting Hill.
\\ \hline
         \mimicplus\ \ftom\ & She is the co - author of Films, Culture, and Popular Culture in Argentina. Her research and writings have appeared or are forthcoming in Latin American Studies, Latin American Studies, Latin American Culture, and other journals.
&He is the author or co - author of many books and many essays on Latin American culture. His main field of study is the history of theater, cinema, and popular culture in Argentina. He was the Editor of Cinemas, Cultures, and Cultures in Argentina,
\\ \hline
         \mimicplus\ \ftom\ & Her work specializes in counseling clients in information security - related incidents, risk management, data collection and compliance and assisting clients with information security - related risks and compliance.
&His work specializes in information security counseling and compliance for clients, dealing with security - related incidents, data - related risk management and software - related compliance incidents.
\\ \hline
         \mimicplus\ \mtof\ & He practices in Fayetteville, Arkansas and has the professional credentials of MD. The NPI Number for Robert L. Firstseine is 1134262688 and he holds a License No. 67 ( Arkansas ).&She practices in Fayetteville, Arkansas and has the professional credentials of RN. The NPI Number for Firstbornee L. Roberts is 1396764048 and she holds a License No. 325 ( Arkansas ).\\ \hline
         \mimicplus\ \mtof\ & Dr. Russ Meka has recently obtained his Ph.D. in Global Affairs at the University of Michigan, Ann Arbor, USA. He has also been a Research Fellow at the Center for Globalization and Society, Bergen University, Norway.
& Ms. Meka has earned her Ph.D. in Global Affairs from Rutgers University, Newark, NJ. She is also an Assistant Professor at Rutgers University, Newark, NJ. Her research interests include global medicine, social justice,
\\ \hline
\end{tabular}
 }
    \caption{Random sample of inverted representations without intervention, alongside an intervention + inversion.}
    \label{tab:random-sample}
    
\end{table*}

\end{document}